\documentclass{article}

\usepackage{times}
\usepackage{booktabs}
\usepackage{graphicx} 
\usepackage{subfigure} 

\usepackage{natbib}

\usepackage{algorithm}
\usepackage{algorithmic}
\usepackage{hyperref}

\usepackage{subfigure} 
\usepackage{float}
\usepackage{amsmath}
\usepackage{color}
\usepackage{array}
\newcolumntype{L}{>{\centering\arraybackslash}m{3cm}}

\begin{document} 

\author{Yotam Hechtlinger, Purvasha Chakravarti \& Jining Qin \\ 
		Department of Statistics \\
        Carnegie Mellon University \\ 
        \{yhechtli,pchakrav,jiningq\}@stat.cmu.edu
}

\title{A Generalization of Convolutional Neural Networks to Graph-Structured Data}
\maketitle

\begin{abstract}
This paper introduces a generalization of Convolutional Neural Networks (CNNs) from low-dimensional grid data, such as images, to graph-structured data. We propose a novel spatial convolution utilizing a random walk to uncover the relations within the input, analogous to the way the standard convolution uses the spatial neighborhood of a pixel on the grid. The convolution has an intuitive interpretation, is efficient and scalable and can also be used on data with varying graph structure. Furthermore, this generalization can be applied to many standard regression or classification problems, by learning the the underlying graph. We empirically demonstrate the performance of the proposed CNN on MNIST, and challenge the state-of-the-art on Merck molecular activity data set.
\end{abstract}

\section{Introduction}

\begin{figure*}
	\centering
   	\includegraphics[width=0.48\textwidth]{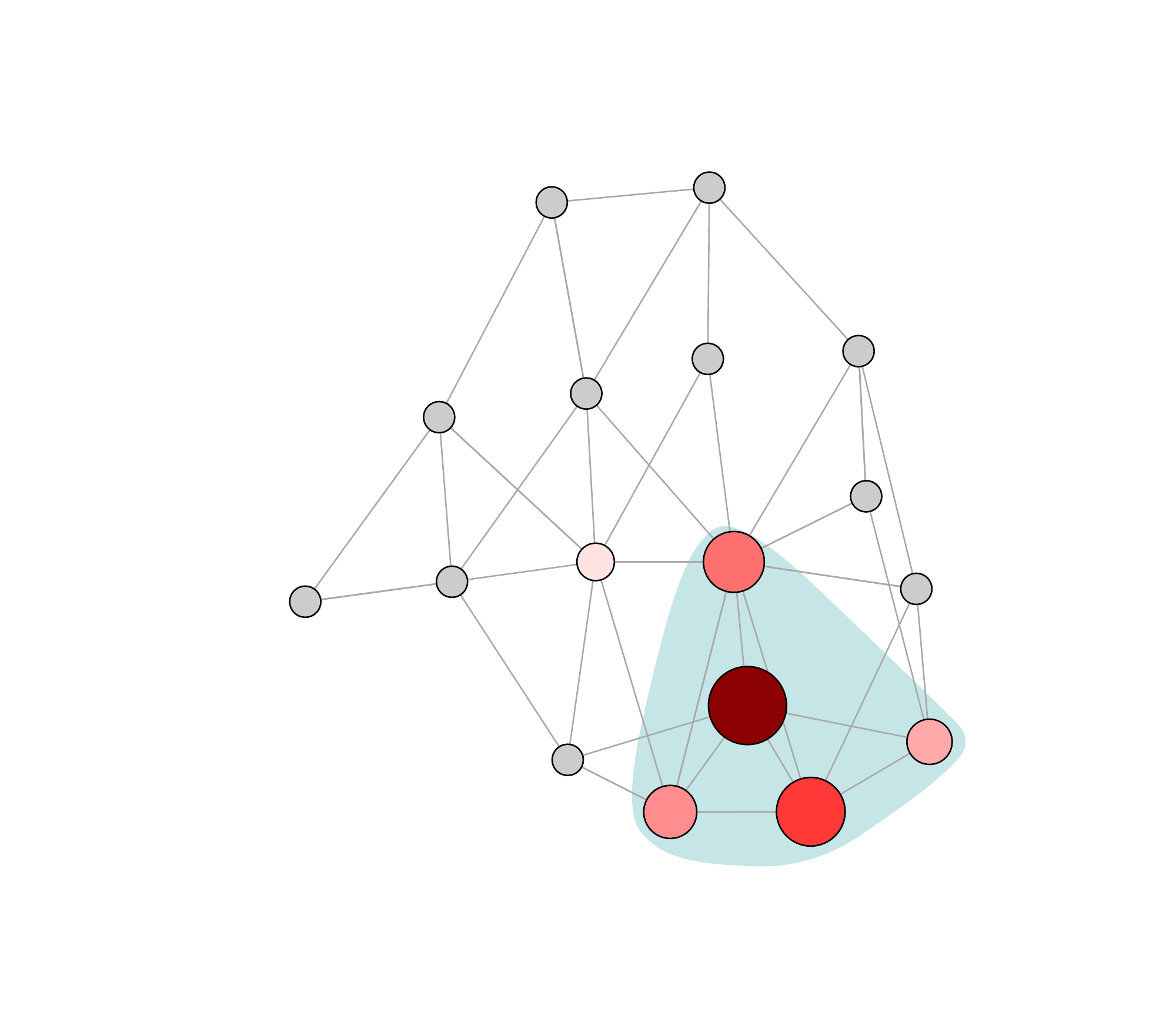}		
   	\includegraphics[width=0.48\textwidth]{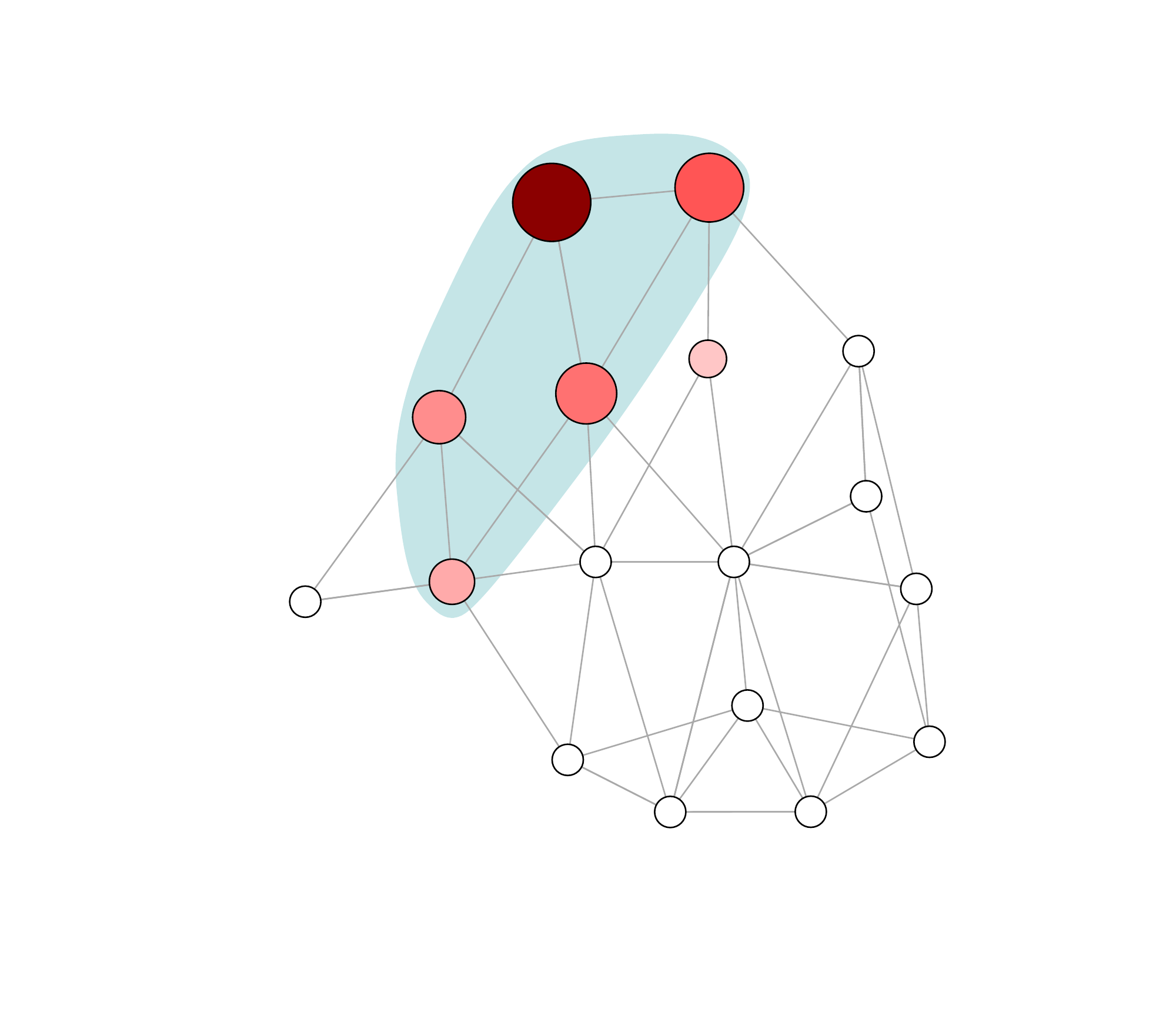}
    \caption{\small Visualization of the graph convolution size $5$. For a given node, the convolution is applied on the node and its $4$ closest neighbors selected by the random walk. As the right figure demonstrates, the random walk can expand further into the graph to higher degree neighbors. The convolution weights are shared according to the neighbors' closeness to the nodes and applied globally on all nodes. }
\label{convolution2}
\end{figure*}

Convolutional Neural Networks (CNNs) are a leading tool used to address a large set of machine learning problems  (\citet{lecun1998gradient}, \citet{lecun2015deep}). They have successfully provided significant improvements in numerous fields, such as image processing, speech recognition, computer vision and pattern recognition, language processing and even the game of Go boards ( \citet{krizhevsky2012imagenet},  \citet{hinton2012deep}, \citet{le2011learning}, \citet{kim2014convolutional}, \citet{silver2016mastering} respectively). 

The major success of CNNs is justly credited to the convolution. But any successful application of the CNNs implicitly capitalizes on the underlying attributes of the input. Specifically, a standard convolution layer can only be applied on grid-structured input, since it learns localized rectangular filters by repeatedly convolving them over multiple patches of the input. Furthermore, for the convolution to be effective, the input needs to be locally connective, which means the signal should be highly correlated in local regions and mostly uncorrelated in global regions. It also requires the input to be stationary in order to make the convolution filters shift-invariant so that they can select local features independent of the spatial location. 

Therefore, CNNs are inherently restricted to a (rich) subset of datasets. Nevertheless, the impressive improvements made by applying CNNs encourage us to generalize CNNs to non-grid structured data that have local connectivity and stationarity properties. The main contribution of this work is a generalization of CNNs to general graph-structured data, directed or undirected, offering a supervised algorithm that incorporates the structural information present in a graph. Moreover our algorithm can be applied to a wide range of regression and classification problems, by first estimating the graph structure of the data and then applying the proposed CNN on it. Active research in learning graph structure from data makes this feasible, as demonstrated by the experiments in the paper.

The fundamental hurdle in generalizing CNNs to graph-structured data is to find a corresponding generalized convolution operator. Recall that the standard convolution operator picks the neighboring pixels of a given pixel and computes the inner product of the weights and these neighbors. We propose a spatial convolution that performs a random walk on the graph in order to select the top $p$ closest neighbors for every node, as Figure \ref{convolution2} shows. Then for each of the nodes, the convolution is computed as the inner product of the weights and the selected $p$ closest neighbors, which are ordered according to their relative position from the node. This allows us to use the same set of weights (shared weights) for the convolution at every node and reflects the dependency between each node and its closest neighbors. When an image is considered as an undirected graph with edges between neighboring pixels, this convolution operation is the same as the standard convolution. 

The proposed convolution possesses many desired advantages:
\begin{itemize}
\item {\bf It is natural and intuitive.} The proposed CNN, similar to the standard CNN, convolves every node with its closest spatial neighbors, providing an intuitive generalization. For example, if we learn the graph structure using the correlation matrix, then selecting a node's $p$ nearest neighbors is similar to selecting its $p$ most correlated variables, and the weights correspond to the neighbors' relative position to the node (i.e. $i^{th}$ weight globally corresponds to the $i^{th}$ most correlated variable for every node). 

\item {\bf It is transferable.} Since the criterion by which the $p$ relevant variables are selected is their relative position to the node, the convolution is invariant to the spatial location of the node on the graph. This enables the application of the same filter globally across the data on all nodes on varying graph structures. It can even be transfered to different data domains, overcoming a known limitation of many other generalizations of CNNs on graphs.

\item {\bf It is scalable.} Each forward call of the graph convolution requires $O\left(N\cdot p\right)$ flops, where $N$ is the number of nodes in the graph or variables. This is also the amount of memory required for the convolution to run. Since $p\ll N$, it provides a scalable and fast operation that can efficiently be implemented on a GPU. 

\item {\bf It is effective.} Experimental results on the Merck molecular activity challenge and the MNIST data sets demonstrates that by learning the graph structure for standard regression or classification problems, a simple application of the graph convolutional neural network gives results that are comparable to state-of-the-art models.
\end{itemize} 

To the best of our knowledge, the proposed graph CNN is the first generalization of convolutions on graphs that demonstrates all of these properties.

\section{Literature review}

Graph theory and differential geometry have been heavily studied in the last few decades, both from mathematical and statistical or computational perspectives, with a large body of algorithms being developed for a variety of problems. This has laid the foundations required for the recent surge of research on generalizing deep learning methods to new geometrical structures. \citet{bronstein2016geometric} provide an extensive review of the newly emerging field. 

Currently, there are two main approaches generalizing CNNs to graph structured data, spectral and spatial approaches (\cite{bronstein2016geometric}). The spectral approach generalizes the convolution operator using the eigenvectors derived from the spectral decomposition of the graph Laplacian. The motivation is to create a convolution operator that commutes with the graph Laplacian similar to the way the regular convolution operator commutes with the Laplacian operator. This approach is studied by \citet{bruna2013spectral} and \citet{henaff2015deep}, which used the eigenvectors of the graph Laplacian to do the convolution, weighting out the distance induced by the similarity matrix. The major drawback of the spectral approach is that it is graph dependent, as it learns filters that are a function of the particular graph Laplacian. This constrains the operation to a fixed graph structure and restricts the transfer of knowledge between different different domains. 

\citet{defferrard2016convolutional} introduce ChebNet, which is a spectral approach with spatial properties. It uses the $k^{th}$ order Chebyshev polynomials of the Laplacian, to learn filters that act on k-hop neighborhoods of the graph, giving them spatial interpretation. Their approach was later simplified and extended to semi-supervised settings by \citet{kipf2016semi}. Although in spirit the spatial property is similar to the one suggested in this paper, since it builds upon the Laplacian, the method is also restricted to a fixed graph structure. 

The spatial approach generalizes the convolution using the graph's spatial structure, capturing the essence of the convolution as an inner product of the parameters with spatially close neighbors. The main challenge with the spatial approach is that it is difficult to find a shift-invariance convolution for non-grid data. Spatial convolutions are usually position dependent and lack meaningful global interpretation. The convolution proposed in this paper is spatial, and utilizes the relative distance between nodes to overcome this difficulty.

Diffusion Convolutional Neural Network (DCNN) proposed by \citet{atwood2016diffusion} is a similar convolution that follows the spatial approach. This convolution also performs a random walk on the graph in order to select spatially close neighbors for the convolution while maintaining the shared weights. DCNN's convolution associates the $i^{th}$ parameter ($w_{i}$) with the $i^{th}$ power of the transition matrix ($P^i$), which is the transition matrix after $i$ steps in a random walk. Therefore, the inner product is considered between the parameters and a weighted average of all the nodes that can be visited in $i$ steps. In practice, for dense graphs the number of nodes visited in $i$ steps can be quite large, which might over-smooth the signal in dense graphs.  Furthermore, \citet{atwood2016diffusion} note that implementation of DCNN requires the power series of the full transition matrix, requiring $O(N^{2})$ complexity, which limits the scalability of the method. 

Another example of a spatial generalization is provided by \citet{bruna2013spectral}, which uses multi-scale clustering to define the network architecture, with the convolutions being defined per cluster without the weight sharing property. \citet{duvenaud2015convolutional} on the other hand, propose a neural network to extract features or molecular fingerprints  from molecules that can be of arbitrary size and shape by designing layers which are local filters applied to all the nodes and their neighbors. 

In addition to the research generalizing convolution on graph, there is active research on the application of different types of Neural Networks on graph structured data. The earliest work in the field is the Graph Neural Network by Scarselli and others, starting with \citet{gori2005new} and fully presented in \citet{scarselli2009graph}. The model connect each node in the graph with its first order neighbors and edges and design the architecture in a recursive way inspired by recursive neural networks. Recently it has been extended by \citet{li2015gated} to output sequences, and there are many other models inspired from the original work on Graph Neural Networks. For example, \citet{battaglia2016interaction} introduce "interaction networks" studying spatial binary relations to learn objects and relations and physics.

The problem of selecting nodes from a graph for a convolution is analogous to the problem of selecting local receptive fields in a general neural network. The work of \citet{coates2011selecting}  suggests selecting the local receptive fields in a feed-forward neural network using the closest neighbors induced by the similarity matrix, with the weights not being shared among the different hidden units.

In contrast to previous research, we suggest a novel scalable convolution operator that captures the local connectivity within the graph and demonstrates the weight sharing property, which helps in transferring it to different domains. We achieve this by considering the closest neighbors, found by using a random walk on the graph, in a way that intuitively extends the spatial nature of the standard convolution. 

\section{Graph Convolutional Neural Network}
The key step which differentiates CNNs on images from regular neural networks is the selection of neighbors on the grid in a $p \times p$ window combined with the shared weight assumption. In order to select the local neighbors of a given node, we use the graph transition matrix and calculate the expected number of visits of a random walk starting from the given node. The convolution for this node, is then applied on the top $p$ nodes with highest expected number of visits from it. In this section, we discuss the application of the convolution in a single layer on a single graph. It is immediate to extend the definition to more complex structures, as will be explicitly explained in section  \ref{ssec:immplementation}. We introduce some notation in order to proceed into further discussion.

\paragraph{Notation:} Let $\mathcal{G}=\left(\mathcal{V},\mathcal{E}\right)$ be a graph over a set of $N$ features, $\mathcal{V}=\left(X_{1},\ldots,X_{N}\right)$, and a set of edges $\mathcal{E}$.  Let $P$ denote the transition matrix of a random walk on the graph, such that $P_{ij}$ is the probability to move from node $X_{i}$ to $X_{j}$. Let the similarity matrix and the correlation matrix of the graph be given by $S$ and $R$ respectively. Define $D$ as a diagonal matrix where $D_{ii} = \sum_{j} S_{ij}$.

\subsection{Transition matrix and expected number of visits}
\subsubsection{Transition matrix existence}
This work assumes the existence of the graph transition matrix $P$. If graph structure of the data is already known, i.e. if the similarity matrix $S$ is already known, then the transition matrix can be obtained, as explained in \citet{lovasz1996random}, by
\begin{equation}
P = D^{-1}S.
\end{equation}
If the graph structure is unknown, it can be learned using several unsupervised or supervised graph learning algorithms. Learning the data graph structure is an active research topic and is not in the scope of this paper. The interested reader can start with \citet{belkin2001laplacian} 
and \citet{henaff2015deep} discussing similarity matrix estimation.  We use the absolute value of the correlation matrix as the similarity matrix, following \citet{roux2008learning} who showed that correlation between the features is usually enough to capture the geometrical structure of images. That is, we assume
\begin{equation}
{S_{ij}} = |R_{ij}| \ \ \forall \ i, j.
\end{equation}

\subsubsection{Expected number of visits}
\begin{figure*}
	\centering
	\includegraphics[width=0.24\textwidth]{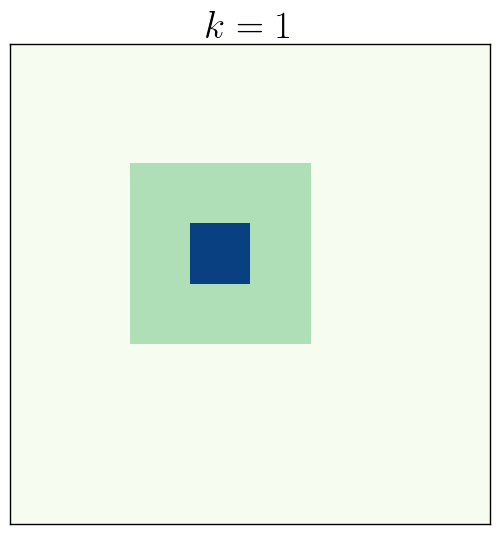}		
   	\includegraphics[width=0.24\textwidth]{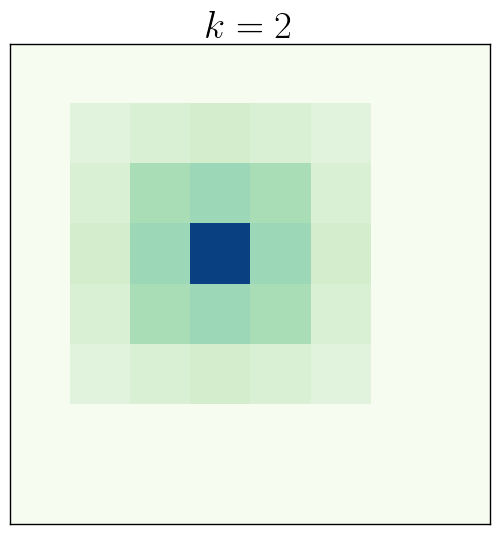}		
   	\includegraphics[width=0.24\textwidth]{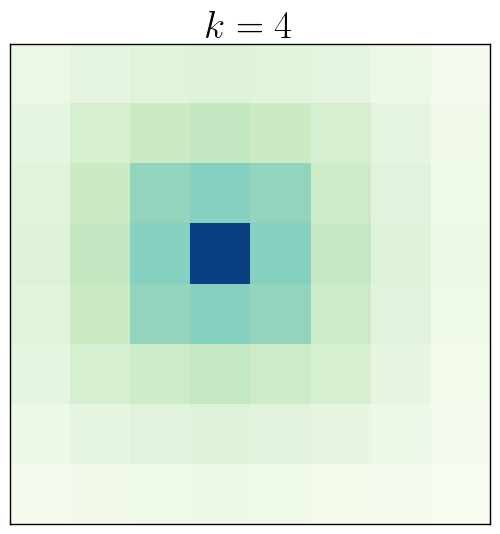}		
   	\includegraphics[width=0.24\textwidth]{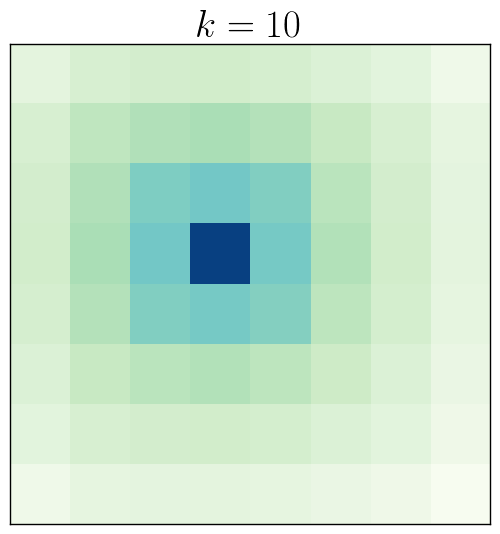}		
	\caption{\small Visualization of a row of $Q^{\left(k\right)}$ on the graph generated over the 2-D grid at a node near the center, when connecting each node to its $8$ adjacent neighbors. For $k=1$, most of the weight is on the node, with smaller weights on the first order neighbors. This corresponds to a standard $3 \times 3$ convolution. As $k$ increases the number of active neighbors also increases, providing greater weight to neighbors farther away, while still keeping the local information.}
\label{convolution}
\end{figure*}

Once we derive the transition matrix $P$, we define $Q^{\left(k\right)}:=\sum_{i=0}^{k}P^{k}$, where $[P^k]_{ij}$ is the probability of transitioning from $X_{i}$ to $X_{j}$ in $k$ steps. 
That is,
\begin{equation}
Q^{\left(0\right)}=I, \ \ Q^{\left(1\right)}=I+P,  \cdots , Q^{\left(k\right)}=\sum_{i=0}^{k}P^{k}.
\end{equation}
Note that $Q^{(k)}_{ij}$ is also the expected number of visits to node $X_{j}$ starting from $X_{i}$ in $k$ steps. The $i^{th}$ row, $Q_{i\cdot}^{(k)}$ provides a measure of similarity between node $X_{i}$ and its neighbors by considering a random walk on the graph. As $k$ increases we incorporate neighbors further away from the node, while the summation gives appropriate weights to the node and its closest neighbors. Figure \ref{convolution} provides a visualization of the matrix $Q$ over the 2-D grid.

\subsection{Convolutions on graphs}

As discussed earlier, each row of $Q^{\left(k\right)}$ can be used to obtain the closest neighbors of a node. Hence, it seems natural to define the convolution over the graph node $X_{i}$  using the $i^{th}$ row of   $Q^{\left(k\right)}$. In order to do so, we denote $\pi_{i}^{(k)}$ as the permutation order of the $i^{th}$ row of $Q^{\left(k\right)}$ in descending order. That is, for every $i = 1, 2, ..., N$  and every $k$,
\[\pi_{i}^{(k)} : \{ 1,2, ..., N\} \longrightarrow  \{ 1,2, ..., N\},\]
such that $Q_{i \pi_{i}^{(k)}(1)} > Q_{i \pi_{i}^{(k)}(2)} > ... > Q_{i \pi_{i}^{(k)}(N)}$.

The notion of ordered position between the nodes is a global feature of all graphs and nodes. Therefore, we can take advantage of it to satisfy the desired shared weights assumption, enabling meaningful and transferable filters. We define $Conv_{1}$ as the size $p$ convolution over the graph $G$ with nodes ${\bf x} = (x_1, \ldots, x_N)^T \in R^{N}$ and weights ${\bf w} \in R^{p}$, for the $p$ nearest neighbors of each node, as the inner product:
\begin{equation}
 Conv_1({\bf x}) =
\left[\begin{array}{ccc}
x_{\pi_{1}^{(k)}(1)} & 
 \cdots & 
x_{\pi_{1}^{(k)}(p)}\\
x_{\pi_{2}^{(k)}(1)} & 
 \cdots & 
x_{\pi_{2}^{(k)}(p)}\\
\vdots & \ddots & \vdots\\
x_{\pi_{N}^{(k)}(1)} & 
 \cdots & 
 x_{\pi_{N}^{(k)}(p)}
\end{array}\right]\cdot\left[\begin{array}{c}
w_{1}\\
w_{2}\\
\vdots\\
w_{p}
\end{array}\right]
\end{equation}

Therefore the weights are decided according to the distance induced by the transition matrix. That is, $w_{1}$ will be convolved with the variable which has the largest value in each row of the matrix $Q^{\left(k\right)}$. For example, when $Q^{\left(1\right)}=I+P$, $w_{1}$ will always correspond to the node itself and $w_{2}$ will correspond to the node's closest neighbor. For higher values of $k$, the order will be determined by the graph's unique structure. 

It should be noted that $Conv_{1}$ doesn't take into account the actual distance between the nodes, and might be susceptible (for example) to the effects of negative correlation between the features. For that reason, we have also experimented with $Conv_{2}$, defined as:

\begin{gather}
 Conv_2({\bf x}) = 
\left[\begin{array}{ccc}
y_{1,\pi_{1}^{(k)}(1)} & 
 \cdots & 
y_{1,\pi_{1}^{(k)}(p)}\\
y_{2,\pi_{2}^{(k)}(1)} & 
 \cdots & 
y_{2,\pi_{2}^{(k)}(p)}\\
\vdots & \ddots & \vdots\\
y_{N,\pi_{N}^{(k)}(1)} & 
 \cdots & 
y_{N,\pi_{N}^{(k)}(p)}
\end{array}\right]\cdot\left[\begin{array}{c}
w_{1}\\
w_{2}\\
\vdots\\
w_{p}
\end{array}\right],\\
\nonumber
\text{ where } \ {\bf x} = \left[\begin{array}{c}
x_{1}\\
x_{2}\\
\vdots\\
x_{N}
\end{array}\right] \ \text{ and } \ y_{i j} = \text{ sign}(R_{ij}) \ Q^{k}_{i j} \ x_j.
\end{gather}
In practice the performance of $Conv_{1}$ was on par with $Conv_{2}$, and the major differences between them were smoothed out during the training process. As $Conv_{1}$ is more intuitive, we decided to focus on using $Conv_{1}$.

\subsection{Selection of the power of \texorpdfstring{$Q$}{TEXT}}
The selection of the value of $k$ is data dependent, but there are two main components affecting its value. Firstly, it is necessary for $k$ to be large enough to detect the top $p$ neighbors of every node. If the transition matrix $P$ is sparse, it might require higher values of $k$. Secondly, from properties of stochastic processes, we know that if we denote $\pi$ as the Markov chain stationary distribution, then
\begin{equation}
\lim_{k\rightarrow\infty}\frac{Q_{ij}^{\left(k\right)}}{k}=\pi_{j} \ \ \forall \ i,j.
\end{equation}
This implies that for large values of $k$, local information will be smoothed out and the convolution will repeatedly be applied on the features with maximum connections. For this reason, we suggest keeping $k$ relatively low (but high enough to capture sufficient features). 

\subsection{Implementation} \label{ssec:immplementation}

\subsubsection{The convolution}
An important feature of the suggested convolution is the complexity of the operation. For a graph with $N$ nodes, a single $p$ level convolution only requires $O(N \cdot p)$ flops and memory, where $p\ll N$. 

Furthermore, similar to standard convolution implementation \citep{chellapilla2006high}, it is possible to represent the graph convolution as a tensor dot product, transferring most of the computational burden to the GPU using highly optimized matrix multiplication libraries.

For every graph convolution layer, we have as an input a $3D$ tensor of $M$ observations with $N$ features at depth $d$. We first extend the input with an additional dimension that includes the top $p$ neighbors of each feature selected by $Q^{\left(k\right)}$, transforming the input dimension from $3D$ to $4D$ tensor as
\[
\left(M, N, d\right)\rightarrow\left(M, N, p, d\right).
\]
Now if we apply a graph convolution layer with $d_{new}$ filters, the convolution weights will be a $3D$ tensor of size $\left(p,d,d_{new}\right)$. Therefore application of a graph convolution which is a tensor dot product between the input and the weights along the $\left(p, d\right)$ axes results in an output of size:
\[
\Big(\left(M, N\right),\left(p, d\right)\Big) \bullet \Big(\left(p, d\right),\left(d_{new}\right)\Big)
=\left(M, N, d_{new}\right).
\]

We have implemented the algorithm using Keras \citep{chollet2015keras} and Theano \citep{2016arXiv160502688short} libraries in Python, inheriting all the tools provided by the libraries to train neural networks, such as dropout regularization, advanced optimizers and efficient initialization methods. The source code is publicly available on Github \footnote{\url{https://github.com/hechtlinger/graph_cnn}}.

\subsubsection{The selection of neighbors}
The major computational effort in this algorithm is the computation of $Q$, which is performed once per graph structure as a pre-processing step. As it is usually a one-time computation, it is not a significant constraint.  

However, for very large graphs, if done naively, this might be challenging. An alternative can be achieved by recalling that $Q$ is only needed in order to calculate the expected number of visits from a given node after $k$ steps in a random walk. In most applications, when the graph is very large, it is also usually very sparse. This facilitates an efficient implementation of Breadth First Search algorithm (BFS). Hence, the selection of the $p$ neighbors can be parallelized and would only require $O(N \cdot p)$ memory for every unique graph structure, making the method scalable for very large graphs, when the number of different graphs is manageable. 

Any problem that has many different large graphs is inherently computationally hard. The graph CNN reduces the memory required after the preprocessing from $O(N^{2})$ to $O(N \cdot p)$ per graph. This is because the only information required from the graph is the $p$ nearest neighbors of every node.

\section{Experiments}

In order to test the feasibility of the proposed CNN on graphs, we conducted experiments on well known data sets functioning as benchmarks: Merck molecular activity challenge and MNIST. These data sets are popular and well-studied challenges in computational biology and computer vision, respectively.

In our implementations, in order to enable better comparisons between the models and reduce the chance of over-fitting during the model selection process, we consider shallow and simple architectures instead of deep and complex ones. The hyper-parameters were chosen arbitrarily when possible rather than being tuned and optimized. Nevertheless, we still report state-of-the-art or competitive results on the data sets.

In this section, we denote a graph convolution layer with $k$ feature maps by $C_k$ and a fully connected layer with $k$ hidden units by $FC_k$.

\subsection{Merck molecular activity challenge}

The Merck molecular activity is a Kaggle \footnote{Challenge website is \url{https://www.kaggle.com/c/MerckActivity}}  challenge which is based on 15 molecular activity data sets. The target is predicting activity levels for different molecules based on the structure between the different atoms in the molecule. This helps in identifying molecules in medicines which hit the intended target and do not cause side effects.

\begin{figure*}
\centering
\begin{tabular}{cc}
      \includegraphics[width = 0.53\textwidth]{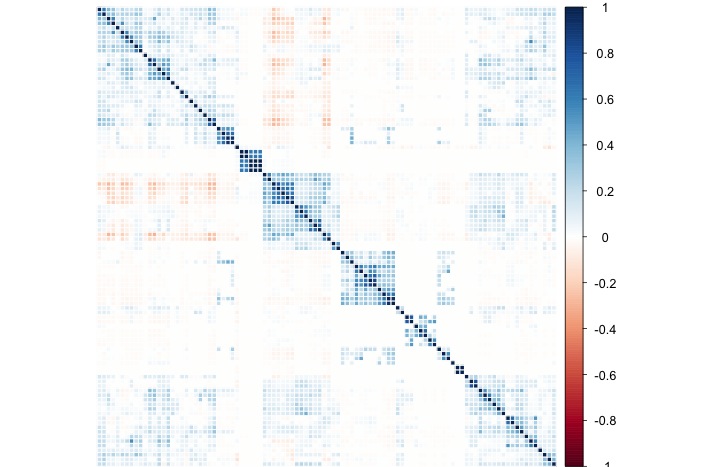}
      \includegraphics[width=0.46\textwidth]{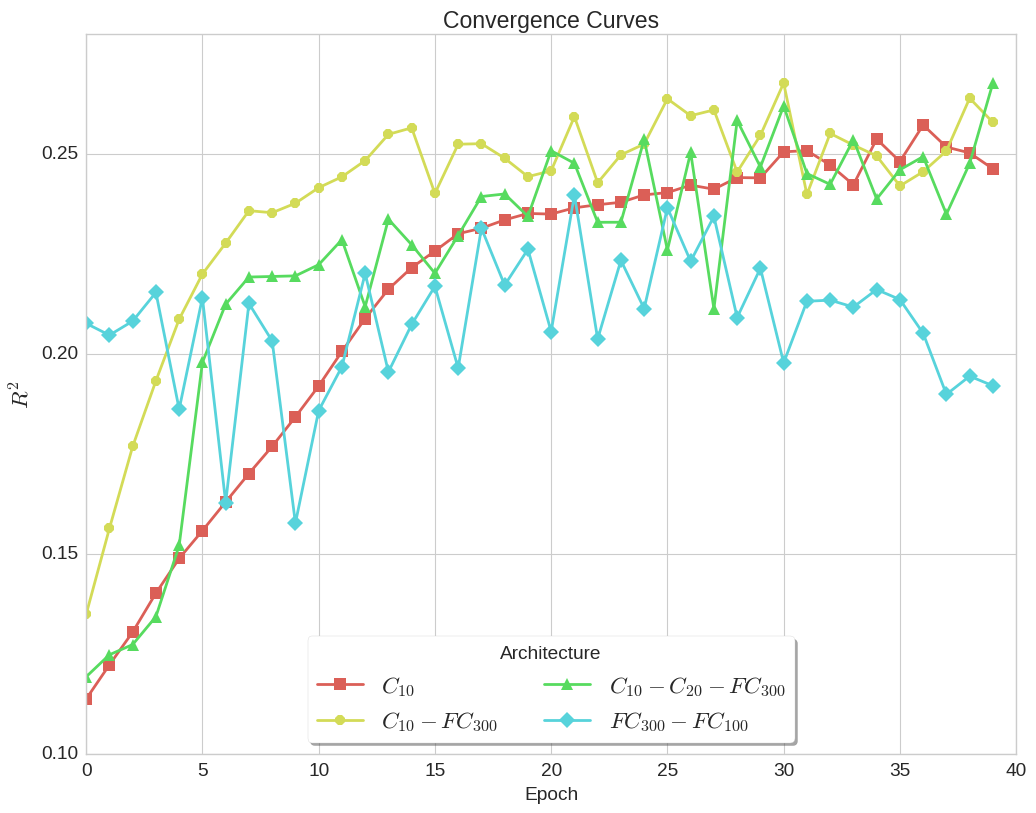}\\
\end{tabular}
\caption{\small \textbf{Left:} Visualization of the correlation matrix between the first 100 molecular descriptors (features) in the DPP4 Merck molecular activity challenge training set. The proposed method utilizes the correlation structure between the features. \textbf{Right:} Convergence of $R^2$ for the different methods on the test set. The graph convolution converges more steadily as it uses fewer parameters.}
\label{molecule}
\end{figure*}

Following \citet{henaff2015deep}, we apply our algorithm on the DPP4 dataset. DPP4 contains $6148$ training and $2045$ test molecules. Some of the features of the molecules are very sparse and are only active in a few molecules. For these features, the correlation estimation is not very accurate. Therefore, we use features that are active in at least $20$ molecules (observations), resulting in $2153$ features. As can be seen in Figure \ref{molecule}, there is significant correlation structure between different features. This implies strong connectivity among the features which is important for the application of the proposed method. 

The training in the experiments was performed using Adam optimization procedure \citep{kingma2014adam} where the gradients are derived by the back-propagation algorithm, using the root mean-squared error loss (RMSE). We used learning rate $\alpha=0.001$, fixed the number of epochs to $40$ and implemented dropout regularization on every layer during the optimization procedure. The absolute values of the correlation matrix were used to learn the graph structure. We found that a small number of nearest neighbors ($p$) between $5$ to $10$ works the best, and used $p=5$ in all models.   

Following the standard set by the Kaggle challenge, results are reported in terms of the squared correlation ($R^2$), that is, 
\[R^2 = \text{Corr}(Y, \hat{Y})^2, \]
where $Y$ is the actual activity level and $\hat{Y}$ is the predicted one. 

The convergence plot given in Figure \ref{molecule} demonstrates convergence of the selected architectures. The contribution of the suggested convolution is explained in view of the alternatives:
\begin{itemize}
\item \textbf{Fully connected Neural Network:} Models first applying convolution followed by a fully connected hidden layer, converge better than more complex fully connected models. Furthermore, convergence in the former methods are more stable in comparison to the fully connected methods, due to the parameter reduction.  
\item \textbf{Linear Regression:} Optimizing over the set of convolutions is often considered as automation of the feature extraction process. From that perspective, a simple application of one layer of convolution, followed by linear regression, significantly outperforms the results of a standalone linear regression. 
\end{itemize}

Table \ref{molecule_error_table} provides more thorough $R^2$ results for the different architectures explored, and compares it to two of the winners of the Kaggle challenge, namely the Deep Neural Network and the random forest in \citet{ma2015deep}. We perform better than both the winners of the Kaggle contest. 

The models in \citet{henaff2015deep} and \citet{bruna2013spectral} use a spectral approach and currently are the state-of-the-art. In comparison to them, we perform better than the Spectral Networks CNN on unsupervised graph structure, which is equivalent to what was done by using the correlation matrix as similarity matrix. The one using Spectral Networks on supervised graph structure holds the state-of-the-art by learning the graph structure. This is a direction we have not yet explored, as graph learning is beyond the scope of this paper, although it will be straightforward to apply the proposed graph CNN in a similar way to any learned graph. 

\begin{table}
\centering
\begin{tabular}{|ccc|}
\toprule
Method & Architecture  & $R^2$ \\
\midrule
OLS Regression  & & 0.135 \\
Random Forest  & & 0.232 \\
{\color{red}Merck winner DNN}& & {\color{red} 0.224}\\
Spectral Networks & C$_{\text{64}}$-P$_{8}$-C$_{64}$-P$_{8}$-FC$_{1000}$ & 0.204 \\ 
{\color{red}Spectral Networks}& {\color{red}C$_{16}$-P$_{4}$-C$_{16}$-P$_{4}$-FC$_{1000}$} & {\color{red} 0.277} \\
 {\color{red}(supervised graph)}  & &  \\
Fully connected NN  & FC$_{300}$-FC$_{100}$ & 0.195 \\
Graph CNN  &C$_{10}$ & { 0.246}\\
Graph CNN  & C$_{10}$-FC$_{100}$ & 0.258 \\
{\color{red} Graph CNN}  & {\color{red} C$_{10}$-C$_{20}$-FC$_{300}$} & {\color{red}0.264}\\
\bottomrule
\end{tabular}
\caption{The squared correlation between the actual activity levels and predicted activity levels, $R^2$ for different methods on DPP4 data set from Merck molecular activity challenge.}
\label{molecule_error_table}
\end{table}

\subsection{MNIST data}

The MNIST data often functions as a benchmark data set to test new machine learning methods. We experimented with two different graph structures for the images. In the first experiment, we considered the images as observations from an undirected graph on the 2-D grid, where each pixel is connected to its $8$ adjoining neighbor pixels. This experiment was done, to demonstrate how the graph convolution compares to standard CNN on data with grid structure. 

We used the convolutions over the grid structure as presented in Figure \ref{convolution} using  $Q^{(3)}$ with $p=25$ as the number of nearest neighbors. Due to the symmetry of the graph, in most regions of the image, multiple pixels are equidistant from the pixel being convolved. In order to solve this, if the ties were broken in a consistent manner, the convolution would be reduced to the regular convolution on a $5 \times 5$ window. The only exceptions to this would be the pixels close to the boundary. To make the example more compelling, we broke ties arbitrarily, making the training process harder compared to regular CNN. Imitating LeNet \cite{lecun1998gradient}, we considered an architecture with  $C_{40}$, $Pooling_{(2\times2)}$, $C_{80}$, $Pooling_{(2\times2)}$ , $FC_{100}$ followed by a linear classifier that resulted in a $0.87\%$ error rate. This is comparable to a regular CNN with the same architecture that achieves an error rate of about $0.75\%$-$0.8\%$. We outperform a fully connected neural network which achieves an error rate of around $1.4\%$, which is expected due to the differences in the complexities of the models. 

In the second experiment, we used the correlation matrix to estimate the graph structure directly from the pixels. Since some of the MNIST pixels are constant (e.g the corners are always black), we restricted the data only to the active $717$ pixels that are not constant. We used $Q^{(1)}$ with $p=6$ as the number of neighbors. This was done in order to ensure that the spatial structure of the image no longer effected the results. With only $6$ neighbors, and a partial subset of the pixels under consideration, the relative location of the top correlated pixels necessarily varies from pixel to pixel. As a result, regular CNNs are no longer applicable on the data whereas the convolution proposed in this paper is. We compared the performance of our CNN to fully connected Neural Networks.

During the training process, we used a  dropout rate of $0.2$ on all layers to prevent over-fitting. In all the architectures the final layer is a standard softmax logistic regression classifier. 

 Table \ref{MNIST_error_table} presents the experimental results. The Graph CNN performs on par with the fully connected neural networks, with fewer parameters. A single layer of graph convolution followed by logistic regression greatly improves the performance of logistic regression, demonstrating the potential of the graph convolution for feature extraction purposes. As with regular convolutions, $C_{20} - FC_{512}$ required over $7$ million parameters as each convolution uses small amount of parameters to generate different maps of the input. This suggests that the graph convolution can be made even more effective with the development of an efficient spatial pooling method on graphs, which is a known but unsolved problem.

\begin{table}
\centering
\begin{tabular}{|ccc|}
\toprule
Method & Error ($\%$) & $\#$ of Parameteres\tabularnewline
\midrule
Logistic Regression & $7.49$ & $7,180$\tabularnewline
$C_{20}$ & $1.94$ & $143,550$\tabularnewline
$C_{20}-C_{20}$ & $1.59$ & $145,970$\tabularnewline
$C_{20}-FC_{512}$ & $1.45$ & $7,347,862$\tabularnewline
$FC_{512}-FC_{512}$ & $1.59$ & $635,402$\tabularnewline
\bottomrule
\end{tabular}
\caption{Error rates of different methods on MNIST digit recognition task without the underlying grid structure.}
\label{MNIST_error_table}

\end{table}

\section{Conclusions}

We propose a generalization of convolutional neural networks from grid-structured data to graph-structured data, a problem that is being actively researched by our community. Our novel contribution is a convolution over a graph that can handle different graph structures as its input. The proposed convolution contains many sought-after attributes; it has a natural and intuitive interpretation, it can be transferred within different domains of knowledge, it is computationally efficient and it is effective. 

Furthermore, the convolution can be applied on standard regression or classification problems by learning the graph structure in the data, using the correlation matrix or other methods. Compared to a fully connected layer, the suggested convolution has significantly fewer parameters while providing stable convergence and comparable performance. Our experimental results on the Merck Molecular Activity data set and MNIST data demonstrate the potential of this approach. 

Convolutional Neural Networks have already revolutionized the fields of computer vision, speech recognition and language processing. We think an important step forward is to extend it to other problems which have an inherent graph structure.

\subsubsection*{Acknowledgments}

We would like to thank Alessandro Rinaldo, Ruslan Salakhutdinov and Matthew Gormley for suggestions, insights and remarks that have greatly improved the quality of this paper. 

\bibliography{reference}
\bibliographystyle{icml2017}

\end{document}